\documentclass{article} 
\usepackage{iclr2027_conference,times}

\usepackage{amsmath,amsfonts,bm}

\def\eqref#1{equation~\ref{#1}}

\def\1{\bm{1}}

\DeclareMathAlphabet{\mathsfit}{\encodingdefault}{\sfdefault}{m}{sl}
\SetMathAlphabet{\mathsfit}{bold}{\encodingdefault}{\sfdefault}{bx}{n}

\usepackage[pagebackref]{hyperref}
\usepackage{url}
\usepackage{amsmath}
\usepackage{amssymb}
\usepackage{booktabs}
\usepackage{graphicx}
\usepackage{algorithm}
\usepackage{algorithmic}
\usepackage{xcolor}
\usepackage{float}
\usepackage{placeins}
\usepackage{caption}
\usepackage{tabularx}
\usepackage{array}
\usepackage{multirow}
\usepackage[capitalise,noabbrev]{cleveref}
\usepackage[most]{tcolorbox}
\usepackage{subcaption}
\usepackage{fontawesome5}
\usepackage{placeins}

\crefname{appendix}{Appendix}{Appendices}
\Crefname{appendix}{Appendix}{Appendices}

\definecolor{mydarkgreen}{rgb}{0,0.45,0.08}
\hypersetup{
    colorlinks=true,
    citecolor=mydarkgreen,
}

\iclrfinalcopy
\author{
Haesung Lee$^{1}$ \quad
Insu Han$^{1}$\thanks{Corresponding author.} \\
$^{1}$KAIST \\
\texttt{\{leehaeseong78,insu.han\}@kaist.ac.kr}
}

\newcommand{\projecturl}{https://haesunglee116.github.io/SACQ_page}
\newcommand{\codeurl}{https://github.com/haesunglee116/SACQ}

\title{SACQ: Structure-Aware Codebook Quantization for Ultra-Low-Bit Vision-Language-Action Models}

\begin{document}

\maketitle
\lhead{Preprint.} 

\begin{abstract}
Vision-Language-Action (VLA) models enable capable robot behavior,
but their large memory footprint limits deployment on resource-constrained robotic platforms.
Weight quantization is a natural remedy, yet post-training quantization (PTQ)
recipes inherited from large language models degrade when applied to VLAs at ultra-low precision.
Existing low-bit PTQ methods often fail to explicitly account for the structure of downstream operations and allocate precision without considering its effect on the policy actions.
We introduce {SACQ}, an ultra-low-bit weight-only vector-codebook quantization framework for VLAs.
SACQ combines
(1) {structure-aware codebook fitting} with role-specific quadratic
metrics that reflect input statistics and downstream sensitivity;
(2) {action-aware mixed-bit allocation} based on the increase in
final-action error as individual layers are compressed; and
(3) {two-stage post-quantization refinement} that reassigns codeword
indices with second-order error compensation and calibrates codebooks and row-group scales against the decoded action.
Using GR00T N1.7, SACQ achieves 93.85\% average success across four LIBERO suites at 1.5 bits per language/action weight while reducing model memory by $4.83\times$. 
With INT8 activation quantization and low-precision attention, it retains 93.55\% success while running $1.51\times$ faster than BF16.
On SimplER, SACQ further achieves 63.76\% and 52.27\% success under Visual Matching and Variant Aggregation at 2 bits per language/action weight, compared to 58.90\% and 53.45\% for BF16, respectively.
\par\medskip
\noindent
\faGlobe\ \href{\projecturl}{\textbf{Project Page:}~\texttt{https://haesunglee116.github.io/SACQ\_page/}} \\
\faGithub\ \href{\codeurl}{\textbf{Code:}~\texttt{https://github.com/haesunglee116/SACQ}}
\end{abstract}
\vspace{-4pt}


\captionsetup[subfigure]{
    justification=centering,
    singlelinecheck=false,
    skip=3pt
}

\begin{figure}[H]
    \centering

    \begin{subfigure}[t]{0.482\linewidth}
        \centering
        \includegraphics[
            width=\linewidth,
            trim=8 12 4 8,
            clip
        ]{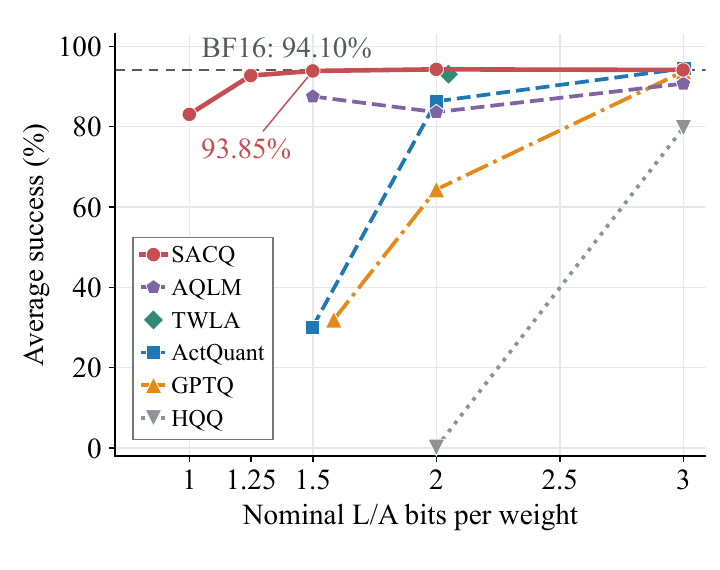}
        \captionsetup{margin={13pt,0pt}}
        \caption{GR00T N1.7 success rate on LIBERO}
        \label{fig:libero_success}
    \end{subfigure}%
    \hspace{0.025\linewidth}%
    \begin{subfigure}[t]{0.482\linewidth}
        \centering
        \includegraphics[
            width=\linewidth,
            trim=6 12 2 8,
            clip
        ]{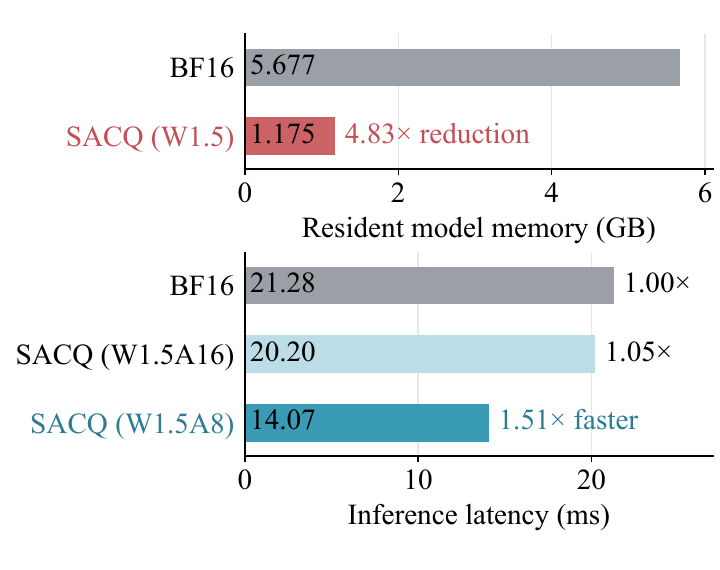}
        \captionsetup{margin={9pt,0pt}}
        \caption{GR00T N1.7 memory and inference efficiency}
        \label{fig:memory_footprint}
    \end{subfigure}

    \caption{
    \textbf{SACQ performance and efficiency on GR00T N1.7.}
    \textbf{(a)} SACQ maintains near-BF16 LIBERO performance down to
    1.5 language/action bits per weight (W1.5), outperforming the baselines
    at lower bit budgets.
    \textbf{(b)} SACQ at W1.5 reduces model memory by $4.83\times$, while W1.5A8 achieves a $1.51\times$ speedup over BF16. A16/A8 denote the 16-/8-bit activation settings, respectively.
    }
    \label{fig:main_result}
\end{figure}

\section{Introduction}

\begin{figure}[t]
    \centering

    \begin{tcolorbox}[
        width=\linewidth,
        colback=white,
        colframe=black!32,
        boxrule=0.6pt,
        arc=2mm,
        outer arc=2mm,
        boxsep=0pt,
        left=5pt,
        right=5pt,
        top=3pt,
        bottom=1pt,
        nobeforeafter
    ]
        \includegraphics[width=\linewidth]{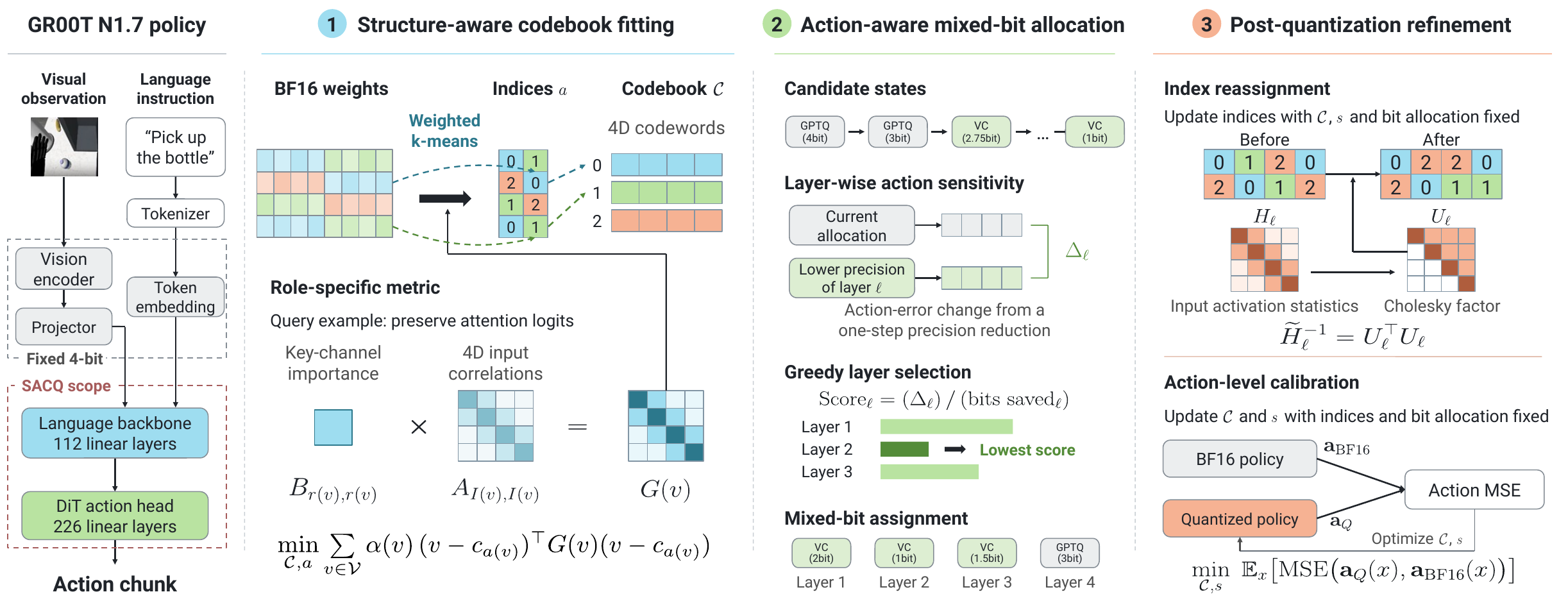}
    \end{tcolorbox}

    \caption{
    \textbf{Overview of SACQ.}
    SACQ proceeds in three stages.
    (1) \textbf{Structure-aware codebook fitting} fits each codebook using a projection-specific metric that reflects both input statistics and downstream sensitivity.
    (2) \textbf{Action-aware mixed-bit allocation} assigns precision according to the increase in final-action error per bit saved.
    (3) \textbf{Two-stage post-quantization refinement} reassigns codeword indices using second-order error compensation, then calibrates the codebooks and row-group scales against the decoded action with the updated indices fixed.
    }
    \label{fig:sacq_overview}
\end{figure}

Vision-Language-Action (VLA) models have made rapid progress in robotic
manipulation by combining visual perception, language understanding, and action
generation within a single policy~\citep{openvla2025,pi05_2025,grootn1_2025}.
Recent models perform strongly across diverse manipulation tasks, establishing
VLAs as an important approach to generalist robot policies.
At the same time, VLA models have grown substantially in scale, creating
significant memory demands for resource-constrained robotic platforms.

Post-training quantization (PTQ) is attractive in this setting because it reduces model memory without requiring full retraining.
However, ultra-low-bit VLA quantization is especially challenging because the
effect of a weight error depends strongly on where it appears in the policy.
For example, errors in query and key projections perturb the attention logits
that determine token mixing, whereas errors in feed-forward projections alter
the hidden representations passed to subsequent operations.
As a result, similar layer-local reconstruction errors can have substantially
different consequences depending on where they occur.
A related challenge is precision allocation: a layer that appears insensitive
under a local reconstruction objective can still have a large effect on the
final action.

Representative PTQ methods such as GPTQ~\citep{gptq2023} minimize layer-wise reconstruction errors using approximate second-order information.
GPTVQ~\citep{gptvq2024} extends this reconstruction-based formulation to vector quantization by incorporating the Hessian of the layer-output reconstruction objective into codeword assignment.
Other vector-quantization methods, including AQLM~\citep{aqlm} and QuIP\#~\citep{quipsharp}, further demonstrate the effectiveness of codebook-based representations for extreme LLM compression.
However, these methods primarily optimize quantization with respect to
layer-local reconstruction and do not explicitly account for the operation
in which each projection output is subsequently used.
For VLAs, recent quantization methods increasingly incorporate properties of action generation into the quantization process.
QVLA~\citep{qvla2026} uses action-related sensitivity to guide mixed-precision allocation, while ActQuant~\citep{actquant2026} combines action-sensitive inter-tensor bit allocation with intra-tensor scale optimization.
HBVLA~\citep{hbvla2026} extends VLA weight quantization to the 1-bit regime, while QuantVLA~\citep{quantvla2026} explicitly considers the structure of the diffusion action transformer in its quantization design.
Despite these advances, jointly accounting for projection-specific downstream sensitivity, final-action error, and ultra-low-bit codebook refinement remains challenging.

To address these challenges, we introduce SACQ, a weight-only vector-codebook framework for ultra-low-bit VLA quantization.
First, {structure-aware codebook fitting} uses projection-specific metrics that reflect each projection's functional role, rather than relying solely on layer-local reconstruction.
Second, {action-aware mixed-bit allocation} assigns precision according
to the increase in final-action error per bit saved when each layer is moved
to a lower-precision state.
Third, {two-stage post-quantization refinement} first reassigns
codeword indices using second-order error compensation and then, with the
updated indices fixed, calibrates codebooks and row-group scales against the
decoded action.
On GR00T N1.7~\citep{grootn1_2025}, SACQ retains 93.85\% average success across the four LIBERO
suites at 1.5 bits per language/action weight, only 0.25
percentage points below the 94.10\% BF16 reference, while reducing resident
model memory from 5.677 GB to 1.175 GB, a 4.83$\times$ reduction.
With an 8-bit activation setting (A8), SACQ achieves a $1.51\times$ speedup over BF16 while retaining 93.55\% average success.
Evaluating GR00T N1.7 on the SimplER Google Robot benchmark~\citep{li24simpler}, SACQ further achieves 63.76\% and 52.27\% success
under Visual Matching and Variant Aggregation, respectively, at 2 bits per language/action weight (bpw), outperforming BF16 on Visual Matching while maintaining comparable performance on Variant Aggregation.

\section{Background}

\subsection{Vision-Language-Action Models}
\label{sec:vla}

A VLA policy maps a visual observation and a language instruction to an action
or a chunk of actions, which is executed on the robot before the next
observation is taken. Its components can be broadly grouped into four modules:
a vision encoder that turns visual inputs into feature representations,
a projector that maps those features into the embedding space of the language
model, a language backbone that processes vision and language jointly, and an
action head that decodes the backbone representation into an executable action.

In both GR00T N1.7 and OpenVLA-OFT, the language backbone consists of decoder
blocks containing the attention projections $q$, $k$, $v$, and $o$, and the
feed-forward projections $\mathrm{gate}$, $\mathrm{up}$, and $\mathrm{down}$.
The action head, however, differs across the two model families.
In GR00T N1.7~\citep{grootn1_2025}, it is a diffusion transformer that
denoises an action chunk over a fixed number of steps; each block contains
attention projections, feed-forward projections, and an adaptive layer
normalization projection that produces per-channel modulation parameters from
the diffusion timestep embedding.
In OpenVLA-OFT~\citep{kim2025fine}, a lightweight MLP-based action head
directly predicts a chunk of continuous-valued actions in a single forward pass.

\subsection{Vector Codebook Quantization}
\label{sec:vc}

Weight quantization replaces each weight matrix $W \in \mathbb{R}^{O \times I}$
by a low-precision representation $\widehat{W}$, reducing the memory needed to
store and load the model. Post-training quantization (PTQ) does this without full model retraining, typically
using a small calibration set drawn from the deployment distribution.

Scalar quantization encodes each weight independently, whereas vector
quantization represents a $d$-dimensional weight vector by a single index
into a shared codebook
$\mathcal{C}=\{c_1,\ldots,c_M\}\subseteq\mathbb{R}^d$, at a nominal index rate of
\begin{equation}
b(M) = \frac{\log_2 M}{d} .
\label{eq:rate}
\end{equation}

We consider a block-scaled vector-codebook representation~\citep{gptvq2024}.
The elements of $W$ are first partitioned into blocks of $g$ consecutive entries, and each block is normalized by its own scale $s$, typically the maximum absolute value within the block, so that codewords are fit to a common range across the matrix. The normalized entries are then grouped into $d$-dimensional
vectors $v$, each of which is assigned an index
$a(v) \in \{1,\ldots,M\}$ and reconstructed in the normalized space as
$\hat v = c_{a(v)}$.
At decoding, each reconstructed vector is rescaled by its corresponding
block scale $s$. Thus, a layer stores its codebook, one index per vector,
and one scale per block.

\subsection{Codebook Optimization}
\label{sec:codebook-opt}

Let $\mathcal{V}$ denote the set of $d$-dimensional vectors covered by one
codebook, obtained from the normalized weights as in \cref{sec:vc}.
Fitting the codebook means choosing the codebook $\mathcal{C}$ and the assignment $a$
that minimize the total distortion over $\mathcal{V}$.
The simplest way to measure that distortion is Euclidean distance,
$\sum_{v} \| v - c_{a(v)} \|^2$, which corresponds to ordinary $k$-means.
This measure treats all $d$ coordinates alike. In general, however, an error
along one coordinate may cost more than an error along another, and the cost
of two coordinates being wrong together may differ from the sum of their
individual costs. Capturing this requires assigning each vector a positive
semidefinite matrix $G(v) \in \mathbb{R}^{d \times d}$ and measuring the error
as a quadratic form, which generalizes the objective to
\begin{equation}
\min_{\mathcal{C}, a}
\sum_{v \in \mathcal{V}}
\alpha(v)
\left(v - c_{a(v)}\right)^\top
G(v)
\left(v - c_{a(v)}\right),
\label{eq:metric-kmeans}
\end{equation}
where $\alpha(v) \ge 0$ is a per-vector weighting factor, and $v$ and $c_{a(v)}$ are points in $\mathbb{R}^d$. Since each normalized vector is rescaled by its block scale $s(v)$ when reconstructed in the original weight space, setting $\alpha(v)=s(v)^2$ accounts for the corresponding block-scale weighting, whereas $\alpha(v)=1$ defines the distortion directly in the scale-normalized space.
The diagonal entries of $G(v)$ state the cost of an error along each of the
$d$ coordinates, and its off-diagonal entries state how errors along different
coordinates combine.
Setting $G(v)=I$ for every $v$ and $\alpha(v)=1$ returns to the Euclidean case above.
Note that \cref{eq:metric-kmeans} assumes that the distortion decomposes into
independent per-vector terms, so that the errors of different weight vectors do not
interact.

The choice of $G$ determines where the codewords are placed.
A codebook covers far more vectors than it has codewords, so most vectors are
represented only approximately, and fitting is necessarily a matter of which
errors to accept and which to avoid.
$G$ is what encodes that ordering.
A common data-aware choice derives $G$ from layer-output reconstruction.
For calibration inputs $X$ and quantization error
$E = W-\widehat{W}$,
it holds $
\left\|XE^\top\right\|_F^2
=
\operatorname{tr}\!\left(EX^\top X E^\top\right)
$.
GPTVQ~\citep{gptvq2024} uses Hessian-derived metrics for
vector assignment.
Given $G(v)$ and $\alpha(v)$, we optimize \cref{eq:metric-kmeans} by alternating between codeword
assignment and codeword updates; the complete updates are provided in
\cref{app:weighted_codebook_optimization}.

Second-order information can also be used to compensate for errors introduced
during sequential quantization. 
GPTQ~\citep{gptq2023} sequentially quantizes weights and uses second-order
information to adjust the remaining unquantized weights so as to reduce the
resulting layer-output error.
GPTVQ~\citep{gptvq2024} extends this principle to vector quantization,
applying the same form of error compensation after quantizing weight vectors.

\section{SACQ: Structure-Aware Codebook Quantization}

SACQ consists of three components: structure-aware codebook fitting,
action-aware mixed-bit allocation, and two-stage post-quantization refinement.
First, we fit each codebook using a metric derived from the downstream
computation in which its projection output is used.
We then allocate precision based on the increase in final-action error as
individual layers are compressed.
Finally, we refine the quantized model by reassigning codeword indices with
second-order error compensation and, with the updated indices fixed, calibrating
the codebooks and row-group scales against the decoded action. An overview of the SACQ pipeline is shown in \cref{fig:sacq_overview}.
We use the vector-codebook representation of \cref{sec:vc} with
$g=64$, $d=4$, and a single codebook per layer; row-group scales are
initialized to the maximum absolute weight value within each group.

\subsection{Structure-Aware Metric Construction}
\label{sec:structure_metric}

The layer-local objective in \cref{sec:codebook-opt} measures
quantization error at the output of each individual projection.
Because its metric is derived from input statistics, the same input-side
weighting is shared across output rows and does not reflect how a particular
output channel is subsequently used.

A projection output, however, is consumed by different downstream operations
depending on its role.
For example, errors in query and key projections perturb the attention logits
that determine token mixing, while errors in value projections are weighted by
the attention probabilities and passed through the output projection.
The output of a gate projection passes through a nonlinearity and is multiplied
elementwise with that of the up projection.
Thus, projection-output errors of similar magnitude can induce substantially
different downstream distortions.

We therefore derive the metric from the distortion of the downstream
computation in which each projection output is used, rather than from the
projection output itself.
The following paragraphs derive this metric for each projection role.

\paragraph{Attention query and key projections.}
The output of a query projection is multiplied with the keys to form the
attention logits. Ignoring the attention scaling factor,
\[
Q = XW_Q^\top,\qquad
K = XW_K^\top,\qquad
L = QK^\top.
\]
Let
$E_Q = W_Q - \widehat{W}_Q$
denote the quantization error of the query projection. For a calibration input
$X$, this changes the queries by
$\Delta Q = XE_Q^\top$.
Holding the keys $K$ fixed, the corresponding change in the attention logits is
$\Delta L = \Delta Q K^\top = X E_Q^\top K^\top$.
The squared Frobenius norm of this logit perturbation is therefore
\begin{equation}
\begin{aligned}
\bigl\| \Delta L \bigr\|_F^2
= \bigl\| X E_Q^\top K^\top \bigr\|_F^2 
= \mathrm{tr}\bigl( E_Q X^\top X E_Q^\top K^\top K \bigr) 
= \mathrm{tr}\bigl( E_Q A E_Q^\top B \bigr),
\end{aligned}
\label{eq:qobj}
\end{equation}
where $A = X^\top X$ and $B = K^\top K$.

Here, $A$ is the Gram matrix of the calibration inputs, capturing the
second-order statistics of the input features.
Similarly, $B$ is the Gram matrix of the keys.
Its diagonal entry $B_{rr}$ measures the energy of key feature $r$, while its
off-diagonal entries capture interactions between quantization errors from
different query output channels.
Compared with the layer-local reconstruction objective in
\cref{sec:codebook-opt}, the output-side weighting $B$ is the new term,
and it arises because the error is measured at the attention logits rather
than at the projection output.

The key projection follows by exchanging the roles of queries and keys. Holding
the queries fixed and expanding in $E_K = W_K - \widehat{W}_K$ gives
$\Delta L = Q E_K X^\top$, with the same input-side Gram $A = X^\top X$ and an
output-side weighting $B = Q^\top Q$.

\paragraph{Separable local objective.}
The loss in \cref{eq:qobj} is not separable across weight vectors. Errors from
different output channels can interact through $B$, and errors on weights
connected to different input features can interact through $A$. The loss
therefore cannot be optimized independently for each weight vector.

To allow independent optimization of each weight vector, we keep only the
diagonal of $B$ and restrict $A$ to the $4\times4$ block corresponding to
its four input coordinates. For a weight vector $v$ belonging to output row $r(v)$ and
occupying input coordinates $I(v)$, this gives
\begin{equation}
G(v) = B_{r(v),\, r(v)} \, A_{I(v),\, I(v)} .
\label{eq:localmetric}
\end{equation}
The block $A_{I(v), I(v)}$ retains correlations among the four input features
within the weight vector, while $B_{r(v), r(v)}$ acts as a scalar weight for the
output channel containing it. Cross-channel terms in $B$ and cross-vector correlations in $A$ are
dropped, yielding a separable, block-diagonal approximation to the full
quadratic objective.
This approximation enables independent codebook fitting, while cross-vector
quantization errors are subsequently addressed by the second-order refinement
in \cref{sec:refinement}.

\paragraph{Attention value and output projections.}
The output of a value projection is not used on its own: it is weighted by the
attention probabilities and then passed through the output projection. Writing
the attention probabilities as $P$,
\[
V = XW_V^\top,\qquad
C = PV,\qquad
Y = CW_O^\top,
\]
where $C$ is the attention context and $Y$ is the block output. Let
$E_V = W_V - \widehat{W}_V$ denote the quantization error of the value
projection. The values change by $\Delta V = X E_V^\top$, and holding $P$ fixed
at the BF16 calibration trajectory, the corresponding change in the block output
is
$\Delta Y = P\Delta V W_O^\top$, or equivalently,
$\Delta Y = PXE_V^\top W_O^\top$,
whose squared Frobenius norm is
\[
\bigl\| \Delta Y \bigr\|_F^2
= \mathrm{tr}\bigl( E_V\, X^\top P^\top P X\, E_V^\top\, W_O^\top W_O \bigr),
\]
of the same form as \cref{eq:qobj}, with $A = X^\top P^\top P X$ and
$B = W_O^\top W_O$. The input-side Gram is thus the input Gram weighted by the
attention probabilities, and the output-side weighting is the Gram of the output
projection.

Applying the same separable approximation, we retain only the diagonals of $P^\top P$ and
of $W_O^\top W_O$. Diagonalizing $P^\top P$ weights each source token by its
squared attention probability and excludes cross terms between different source
tokens. For source token $s$, that diagonal entry is the sum over all attention
heads $h$ and query positions $t$,
\[
\omega_s = \sum_{h,t} P_{h,t,s}^2,
\]
which measures how often and how strongly token $s$ is attended to.
Diagonalizing $W_O^\top W_O$ gives output channel $r$ the weight
$\| W_O[:,r] \|_2^2$ and excludes coupling between channels. The input-side Gram
and output-side weighting are therefore
\begin{equation}
A = \mathbb{E}\Bigl[ \sum_s \omega_s\, x_s x_s^\top \Bigr],
\qquad
B_{rr} = \bigl\| W_O[:,r] \bigr\|_2^2 .
\label{eq:vmetric}
\end{equation}
The input to an output projection is the context $C$ itself, so its input-side
Gram is $A = \mathbb{E}[C^\top C]$. We do not model a downstream operator for
its output and target the reconstruction of that output directly, so $B = I$.

\paragraph{Feed-forward projections.}
Let $h$ denote the input activation to the feed-forward block, which is given by
\[
g = W_g h,\qquad
u = W_u h,\qquad
z = \operatorname{SiLU}(g)\odot u,\qquad
y = W_d z.
\]

For the down projection, whose input is the post-activation representation $z$, we use
$A=\mathbb{E}[zz^\top]$ and $B=I$.
The gate and up projections interact nonlinearly through SwiGLU before
passing through the downstream projection $W_d$. For a four-dimensional
weight vector $v$ in output channel $r$, let $h_v$ denote the input
activations corresponding to its four weights, and let $g_r$ and $u_r$
denote the corresponding gate and up activations. We use the diagonal metrics
\begin{equation}
\begin{aligned}
G_{\mathrm{gate}}(v)
&=
\|W_d[:,r]\|_2^2
\operatorname{Diag}\!\left(
\mathbb{E}\!\left[
\bigl(\operatorname{SiLU}'(g_r)u_r\bigr)^2
(h_v\odot h_v)
\right]
\right),\\
G_{\mathrm{up}}(v)
&=
\|W_d[:,r]\|_2^2
\operatorname{Diag}\!\left(
\mathbb{E}\!\left[
\operatorname{SiLU}(g_r)^2
(h_v\odot h_v)
\right]
\right).
\end{aligned}
\label{eq:ffn_metrics}
\end{equation}
The full first-order derivation is provided in
\cref{app:backbone_metrics}.

\paragraph{Action-head projections.}
The DiT action head follows the same structure-aware principle, while
accounting for its model-specific components: separate query- and
key/value-side input statistics in cross-attention, GELU-based
feed-forward sensitivities, and output-side weighting for AdaLN
modulation. The complete role-specific metrics and derivations are
provided in \cref{app:action_head_metrics}.

\paragraph{Fitting.}
With $G(v)$ determined for each projection role, we apply the
metric-weighted codebook optimization described in
\cref{sec:codebook-opt}.
In SACQ, we set $\alpha(v)=1$ and perform codebook fitting directly in the scale-normalized weight space, rather than additionally weighting each vector by $s(v)^2$. 
For each layer, we fit the codebook on a sampled subset of vectors and then,
with the codebook fixed, recompute assignments over all vectors in the layer.
Further fitting details are provided in
\cref{app:fitting_configuration}.

\subsection{Action-Aware Mixed-Bit Allocation}
\label{sec:mixed_bit_allocation}

The codebook fitting in \cref{sec:structure_metric} is performed independently for each layer and does
not determine how many bits each layer receives. This section determines that
assignment under a global storage budget. Each candidate layer takes one of a set of states ordered by precision,
\[
\mathrm{GPTQ\,(4bit)}
\rightarrow
\mathrm{GPTQ\,(3bit)}
\rightarrow
\mathrm{VC\,(2.75bit)}
\rightarrow
\mathrm{VC\,(2.5bit)}
\rightarrow
\cdots
\rightarrow
\mathrm{VC\,(1bit)}.
\]
Here, $\mathrm{VC}\,(r\mathrm{bit})$ denotes a vector-codebook state
with nominal rate $r$ bits per weight, whose codebook size follows
from \cref{eq:rate} as $M=2^{dr}$.
For the 1-bit operating point, we additionally allow the
$\mathrm{VC\,(0.75bit)}$ state.
Only transitions between adjacent states are allowed.
Within these states, precision is allocated according to the effect on the final decoded action rather than
the layer-wise reconstruction error. For each layer $\ell$, we measure the increase $\Delta_\ell$ in final-action
relative error caused by moving it one state down, averaged over the calibration
set with equal weighting across suites. This is measured by
actually running the model with that layer at lower precision, rather than
relying on a layer-local sensitivity proxy.

For allocation, we compute the calculated bits per weight of the language backbone and action
head as the total stored bits of each component divided by its parameter count,
including metadata such as codebooks and row-group scales. This
overhead is not negligible at low rates, so the allocation is driven by this
quantity rather than by the nominal rate. We write
$\mathrm{cost}_\ell(\rho)$ for the storage cost of layer $\ell$ in state $\rho$.
At each step, we select the component with the higher calculated bits per
weight and, within it, choose the layer with the smallest increase in action
error per bit saved,
\begin{equation}
\ell^\star
=
\mathop{\textrm{argmin}}_{\ell}
\frac{\Delta_\ell}
{\mathrm{cost}_\ell(\rho_\ell)
-
\mathrm{cost}_\ell(\rho'_\ell)},
\label{eq:allocation_rule}
\end{equation}
where $\rho_\ell$ is the current state of layer $\ell$ and $\rho'_\ell$ is its next
lower-precision state. This component-balancing rule prevents either the
language backbone or the action head from being compressed disproportionately
relative to the other.
Further details of the allocation procedure are provided in
\cref{app:allocation}.

\subsection{Post-Quantization Refinement}
\label{sec:refinement}

After structure-aware fitting and mixed-bit allocation, we refine the
quantized policy in two stages. We first reassign the discrete codeword
indices using second-order compensation and then, with the updated indices
fixed, calibrate the codebooks and row-group scales against the decoded action.
Both refinement stages are applied only to vector-codebook layers. Layers remaining in GPTQ states are kept fixed.

\paragraph{Index reassignment.}
Following the second-order compensation mentioned in
\cref{sec:codebook-opt}, we refine the discrete assignments
obtained from structure-aware fitting.
Structure-aware codebook fitting and index reassignment serve different roles:
$G(v)$ guides projection-specific codebook fitting, while index reassignment
compensates cross-vector quantization errors.
For each target layer, we run the frozen quantized policy on the calibration
data and collect the input second-moment matrix
\[
H_\ell
=
\frac{1}{N_\ell}
\sum_{t=1}^{N_\ell} x_t x_t^\top .
\]
Here, $x_t$ denotes the $t$-th input activation vector collected for layer $\ell$,
and $N_\ell$ is the total number of such input activation vectors.
Although $H_\ell$ is collected from the quantized policy, the reconstruction
target for reassignment is the original BF16 weight.
Let $\widetilde H_\ell$ denote the damped version of $H_\ell$, and write
$\widetilde H_\ell^{-1}=U_\ell^\top U_\ell$,
where $U_\ell$ is the upper-triangular Cholesky factor.
Let $J$ denote the coordinates of the current $d$-dimensional vector and $R$
the remaining coordinates, with $U_{JJ}$ and $U_{JR}$ denoting the
corresponding blocks of $U_\ell$.
We use $\bar w$ for the current reconstruction target after compensation from
previously processed vectors, and $q_k$ for the candidate reconstruction using
codeword $k$ under the fixed row-group scale.
The current vector is first reassigned, and the resulting residual is
then propagated to the remaining coordinates:
\begin{equation}
\begin{aligned}
a_J^*
&=
\arg\min_k
\left\|
(\bar{w}_J-q_k)U_{JJ}^{-1}
\right\|_2^2, \\
\bar{w}_R
&\leftarrow
\bar{w}_R
-
(\bar{w}_J-q_{a_J^*})
U_{JJ}^{-1}U_{JR}.
\end{aligned}
\label{eq:index_reassignment}
\end{equation}

Subsequent vectors are reassigned using this compensated target, allowing
their assignments to absorb the preceding quantization error.
Codebooks, row-group scales, and the mixed-bit allocation remain fixed during
this stage; only the codeword indices are updated.

\paragraph{Action-level calibration.}
After index reassignment, we freeze the updated codeword indices $a^*$ and
optimize only the codebooks and row-group scales to match the BF16 action.
Following prior work on codebook fine-tuning~\citep{aqlm}, we denote
the codebooks and row-group scales by $\mathcal{C}$ and $s$, respectively.
For a calibration sample $x$ and fixed $a^*$, we minimize

\begin{equation}
\min_{\mathcal{C},s}
\mathbb{E}_x
\left[
\frac{1}{N_x}
\left\|
\mathbf{a}_Q(x;\mathcal{C},s,a^\ast)
-
\mathbf{a}_{\mathrm{BF16}}(x)
\right\|_F^2
\right],
\label{eq:action_calibration}
\end{equation}

where $N_x$ is the number of elements in the action tensor.
This directly calibrates the codebooks and scales so that the quantized
policy matches the BF16 action.
Keeping the indices fixed preserves the storage rate during calibration.
Further optimization and validation details are provided in \cref{app:action_calibration}.
\section{Experiments}
\label{sec:experiments}

\subsection{Experimental Setup}
\label{sec:experimental_setup}

We evaluate GR00T N1.7~\citep{grootn1_2025}, quantizing 338 linear layers (112 language-backbone and 226 DiT action-head layers) while keeping the remaining 128 linear layers and token embedding at 4-bit. Activations remain BF16, with BF16 serving as the full-precision reference.
We compare against HQQ~\citep{hqq2023},
GPTQ~\citep{gptq2023}, ActQuant~\citep{actquant2026}, 
TWLA~\citep{twla2026} and 
AQLM~\citep{aqlm}
under the same language/action quantization
scope.
Calibration-based methods use the same calibration data as SACQ and follow their respective quantization and optimization procedures.
We evaluate on LIBERO~\citep{libero2023} (Spatial, Object, Goal, and Long) and SimplER~\citep{li24simpler} (Visual Matching and Variant Aggregation). Full evaluation and quantization settings are provided in
\cref{app:experimental_details}, with baseline implementation
details in \cref{app:baseline_details}.

\begin{table}[t]
    \centering
    \captionsetup{skip=6pt}
    \caption{
    \textbf{Comparison of SACQ with existing quantization methods on GR00T N1.7
    across the four LIBERO suites.}
    We report success rates (\%) against HQQ, GPTQ, ActQuant, TWLA, and AQLM at
    comparable language/action bit budgets.
    L/A bpw denotes the parameter-weighted average bits per weight over the
    338 language-backbone and action-head candidate layers.
    }
    \label{tab:libero_main}

    \small
    \renewcommand{\arraystretch}{1.08}

    \begin{tabular*}{\linewidth}{@{\extracolsep{\fill}}lccccccc@{}}
        \toprule
        Method & L/A bpw & Spatial & Object & Goal & Long & Avg. & Memory (GB) \\
        \midrule

        BF16
        & 16.00 & 94.2 & 97.6 & 95.2 & 89.4 & 94.10 & 5.677 \\

        \midrule
        HQQ
        & 3.00 & 75.0 & 93.6 & 76.4 & 73.8 & 79.70 & 1.660 \\

        GPTQ
        & 3.00 & \textbf{95.2} & \textbf{97.4} & \textbf{96.6}
        & 85.6 & 93.70 & 1.982 \\

        ActQuant
        & 3.00 & 94.8 & 96.2 & 96.4
        & \textbf{90.4} & \textbf{94.45} & 1.562 \\

        AQLM
        & 3.00 & 93.6 & \textbf{97.4} & 92.4
        & 79.4 & 90.70 & 1.551 \\

        \textbf{SACQ}
        & 3.00 & 94.8 & 96.8 & \textbf{96.6}
        & 88.2 & 94.10 & 1.902 \\

        \midrule
        HQQ
        & 2.00 & 0.0 & 0.0 & 0.4 & 0.0 & 0.10 & 1.368 \\

        GPTQ
        & 2.00 & 57.2 & 90.2 & 49.8 & 60.2 & 64.35 & 1.528 \\

        ActQuant
        & 2.00 & 92.2 & 96.6 & 88.6 & 67.8 & 86.30 & 1.306 \\

        TWLA
        & 2.05 & 93.6 & \textbf{97.4} & 96.4 & 84.8 & 93.05 & 2.772 \\

        AQLM
        & 2.00 & 93.0 & 95.0 & 76.4
        & 70.0 & 83.60 & 1.599 \\

        \textbf{SACQ}
        & 2.00 & \textbf{93.8} & 96.6 & \textbf{97.0}
        & \textbf{89.6} & \textbf{94.25} & 1.310 \\

        \midrule
        GPTQ-Ternary
        & 1.58 & 39.4 & 44.4 & 10.6 & 33.4 & 31.95 & 1.337 \\

        ActQuant
        & 1.50 & 48.4 & 35.4 & 26.6 & 9.8 & 30.05 & 1.169 \\

        AQLM
        & 1.50 & 89.8 & 96.4 & 88.0
        & 75.8 & 87.50 & 1.138 \\

        \textbf{SACQ}
        & 1.50 & \textbf{93.2} & \textbf{97.0} & \textbf{96.2}
        & \textbf{89.0} & \textbf{93.85} & 1.175 \\

        \midrule
        \textbf{SACQ}
        & 1.25 & 92.0 & 96.8 & 94.4 & 87.6 & 92.70 & 1.114 \\

        \midrule
        \textbf{SACQ}
        & 1.00 & 86.8 & 93.6 & 86.2 & 65.6 & 83.05 & 1.069 \\

        \bottomrule
    \end{tabular*}
\end{table}

\begin{table}[t]
    \centering
    \captionsetup{skip=6pt}
    \caption{
    \textbf{Comparison of SACQ with existing quantization methods on the
    SimplER Google Robot benchmark using GR00T N1.7.}
    We report success rates (\%) under the Visual Matching (VM) and
    Variant Aggregation (VA) protocols against GPTQ, ActQuant, TWLA, and AQLM
    at comparable language/action bit budgets.
    L/A bpw denotes the parameter-weighted average bits per weight over the
    338 language-backbone and action-head candidate layers.
    }
    \label{tab:simpler_main}

    \small
    \renewcommand{\arraystretch}{1.08}

    \begin{tabular*}{\linewidth}{@{\extracolsep{\fill}}lccccccc@{}}
        \toprule
        Method & L/A bpw & Pick & Move & Drawer & Apple & Avg. & Memory (GB) \\
        \midrule

        \multicolumn{8}{c}{\textit{Visual Matching (VM)}} \\
        \midrule

        BF16
        & 16.00 & 86.00 & 91.25 & 45.37 & 12.96 & 58.90 & 5.677 \\

        GPTQ
        & 2.00 & 70.33 & 25.00 & 6.94 & 0.00 & 25.57 & 1.524 \\

        ActQuant
        & 2.00 & 90.67 & 62.92 & 32.87 & 0.00 & 46.61 & 1.331 \\

        TWLA
        & 2.06 & 84.00 & 72.08 & 47.22 & 7.41 & 52.68 & 2.782 \\

        AQLM
        & 2.00 & 73.33 & 26.25 & 24.54 & 0.00 & 31.03 & 1.595 \\

        \textbf{SACQ}
        & 2.00
        & \textbf{91.33}
        & \textbf{89.17}
        & \textbf{49.54}
        & \textbf{25.00}
        & \textbf{63.76}
        & 1.342 \\

        \textbf{SACQ}
        & 1.50 & 84.67 & 73.75 & 38.89 & 31.48 & 57.20 & 1.191 \\

        \midrule
        \multicolumn{8}{c}{\textit{Variant Aggregation (VA)}} \\
        \midrule

        BF16
        & 16.00 & 83.27 & 81.83 & 21.69 & 26.98 & 53.45 & 5.677 \\

        GPTQ
        & 2.00 & 67.88 & 23.00 & 2.12 & 0.00 & 23.25 & 1.524 \\

        ActQuant
        & 2.00 & 76.97 & 64.33 & 11.38 & 2.12 & 38.70 & 1.331 \\

        TWLA
        & 2.06 & 79.88 & 68.83 & 14.29 & 29.63 & 48.16 & 2.782 \\

        AQLM
        & 2.00 & 65.82 & 32.50 & 5.29 & 0.00 & 25.90 & 1.595 \\

        \textbf{SACQ}
        & 2.00
        & \textbf{80.73}
        & \textbf{77.83}
        & \textbf{17.72}
        & \textbf{32.80}
        & \textbf{52.27}
        & 1.342 \\

        \textbf{SACQ}
        & 1.50 & 75.39 & 74.17 & 4.76 & 24.34 & 44.67 & 1.191 \\

        \bottomrule
    \end{tabular*}
\end{table}

\subsection{Main Results}

\textbf{LIBERO.}
\cref{tab:libero_main} compares SACQ with existing quantization methods
across the four LIBERO suites. SACQ achieves 93.85\% average success at
1.5 L/A bpw, only 0.25 percentage points below the 94.10\% BF16
reference, while reaching 94.25\% and 94.10\% at 2 and
3 L/A bpw, respectively. 
For comparison, TWLA reaches 93.05\% and AQLM 83.60\% at 2 L/A bpw,
while AQLM reaches 87.50\% at 1.5 L/A bpw.
This low-bit performance also translates into a substantial reduction
in model footprint. At 1.5 L/A bpw, SACQ reduces model
memory from 5.677 GB to 1.175 GB, a 4.83$\times$ reduction, while
remaining within 0.25 percentage points of BF16.

\textbf{SimplER.}
\cref{tab:simpler_main} compares SACQ with existing quantization methods
on the SimplER Google Robot benchmark. At 2 L/A bpw,
SACQ achieves average success rates of 63.76\% on VM and 52.27\% on VA,
compared with the BF16 references of 58.90\% and 53.45\%, respectively.
SACQ also achieves higher success than the evaluated quantized baselines
under both protocols. Further reducing SACQ to 1.5 L/A bpw decreases
success to 57.20\% on VM and 44.67\% on VA.

\begin{table}[t]
    \centering
    \captionsetup{skip=6pt}
    \caption{
    \textbf{Inference efficiency of SACQ on GR00T N1.7 across the four LIBERO suites.}
    We report task success rates (\%), whole-model inference latency, and relative speed with respect to BF16.
    }
    \label{tab:inference_efficiency}

    \small
    \renewcommand{\arraystretch}{1.08}

    \begin{tabular*}{\linewidth}{@{\extracolsep{\fill}}lccccccc@{}}
        \toprule
        Configuration & Spatial & Object & Goal & Long & Avg. & Latency (ms) & Relative speed \\
        \midrule

        BF16
        & 94.2 & 97.6 & 95.2 & 89.4 & 94.10 & 21.28 & 1.00$\times$ \\

        SACQ (W1.5A16)
        & 93.2 & 97.0 & 96.2 & 89.0 & 93.85 & 20.20 & 1.05$\times$ \\

        \textbf{SACQ (W1.5A8)}
        & 93.8 & 96.2 & 95.4 & 88.8 & 93.55
        & \textbf{14.07} & \textbf{1.51$\times$} \\

        \bottomrule
    \end{tabular*}
\end{table}

\subsection{Inference Efficiency}
\label{sec:inference_efficiency}

At 1.5 L/A bpw (W1.5), we evaluate two activation-precision settings: a 16-bit activation setting (W1.5A16) and an 8-bit activation setting (W1.5A8). In A8, linear projections use dynamic per-row symmetric INT8 activation quantization with INT8 GEMMs, while action self-attention and vision attention use an INT8-QK/FP8-PV attention path. The attention cores in action cross-attention and language attention remain in BF16. The runtime additionally uses optimized packed-weight decoding kernels and CUDA Graph execution while retaining SACQ's compressed weight representation. For a fair comparison, BF16 uses the same common runtime optimizations and CUDA Graph execution.

As shown in \cref{tab:inference_efficiency}, W1.5A8 achieves 93.55\% average success on LIBERO, only 0.30 percentage points below W1.5A16 and 0.55 percentage points below BF16, while achieving a $1.51\times$ speedup over BF16. Using the official GR00T inference path, we measure whole-model inference latency on an RTX 4090 with batch size 1 and four denoising steps. Wall-clock time is measured with GPU synchronization before and after each model call, and we report the median of 20 measurements after five warmup iterations, excluding preprocessing.

\begin{table}[t]
    \centering
    \captionsetup{skip=6pt}
    \caption{
    \textbf{Component and structure-aware ablations of SACQ on LIBERO.}
    We evaluate the three components at 1.5 L/A bpw and
    structure-aware fitting at 1.25 L/A bpw.
    }
    \label{tab:component_ablation}

    \small
    \setlength{\tabcolsep}{1.0pt}
    \renewcommand{\arraystretch}{1.08}

    \begin{tabular*}{\linewidth}{
        @{\extracolsep{\fill}}
        lccccccccc
        @{}
    }
        \toprule
        Setting
        & L/A bpw
        & Alloc.
        & Struct.
        & Refine.
        & Spatial
        & Object
        & Goal
        & Long
        & Avg. \\
        \midrule

        BF16
        & 16.0
        & -- & -- & --
        & 94.2 & 97.6 & 95.2 & 89.4 & 94.10 \\

        \midrule

        Uniform Euclidean VQ
        & 1.50
        & -- & -- & --
        & 0.0 & 0.0 & 0.0 & 0.0 & 0.00 \\

        \hspace{0.6em}+ Mixed-bit allocation
        & 1.50
        & \checkmark & -- & --
        & 0.0 & 0.0 & 0.0 & 0.0 & 0.00 \\

        \hspace{0.6em}+ Structure-aware fitting
        & 1.50
        & \checkmark & \checkmark & --
        & 44.8 & 73.4 & 31.4 & 40.4 & 47.50 \\

        \textbf{Full SACQ}
        & 1.50
        & \checkmark & \checkmark & \checkmark
        & 93.2 & 97.0 & 96.2 & 89.0 & 93.85 \\

        \midrule

        w/o Struct. (Euclidean)
        & 1.50
        & \checkmark & -- & \checkmark
        & 95.2 & 95.8 & 93.8 & 85.6 & 92.60 \\

        w/o Allocation (Uniform)
        & 1.50
        & -- & \checkmark & \checkmark
        & 94.2 & 96.8 & 95.8 & 86.2 & 93.25 \\

        \midrule

        w/o Struct. (Euclidean)
        & 1.25
        & \checkmark & -- & \checkmark
        & 91.8 & 94.4 & 91.2 & 73.2 & 87.65 \\

        \textbf{Full SACQ}
        & 1.25
        & \checkmark & \checkmark & \checkmark
        & 92.0 & 96.8 & 94.4 & 87.6 & 92.70 \\

        \bottomrule
    \end{tabular*}
\end{table}

\section{Analysis and Ablations}
\label{sec:analysis}

We analyze each SACQ component and structure-aware fitting under aggressive compression using the LIBERO protocol of \cref{sec:experiments}.
At 1.5 L/A bpw, Euclidean VQ with mixed-bit allocation
fails to preserve the policy, whereas structure-aware fitting raises
the pre-refinement average success to 47.50\%. Post-quantization
refinement then recovers the full SACQ model to 93.85\%. Removing
structure-aware fitting or mixed-bit allocation reduces the final
success to 92.60\% and 93.25\%, respectively. 
At 1.25 L/A bpw, structure-aware fitting improves the
average success from 87.65\% to 92.70\%, with the largest gain on
LIBERO-Long from 73.2\% to 87.6\% (+14.4 percentage points).
Additional analysis under limited post-quantization refinement is provided in
\cref{app:limited_refinement}.

\section{Conclusion}

We introduced SACQ for ultra-low-bit VLA quantization.
SACQ combines structure-aware codebook fitting, which accounts for
the functional role of each projection, action-aware mixed-bit
allocation, which assigns precision according to its effect on the final
decoded action, and two-stage post-quantization refinement for
recovering the quantized policy.

On GR00T N1.7, SACQ maintains performance close to the BF16 reference on
LIBERO even at 1.5 L/A bpw, while reducing resident model memory from
5.677 GB to 1.175 GB, a 4.83$\times$ reduction.
With the 8-bit activation setting (A8), SACQ at 1.5 L/A bpw retains 93.55\% average success while achieving a $1.51\times$ inference speedup over BF16.
SACQ also maintains strong performance on SimplER at 2 L/A bpw.
Our analysis further shows that structure-aware fitting becomes particularly
effective under more aggressive compression and limited post-quantization
refinement (\cref{app:limited_refinement}).

Additional experiments demonstrate that this behavior extends beyond the
main setting. On OpenVLA-OFT evaluated on LIBERO, SACQ achieves 94.05\%
average success at 2 V/L bpw, compared with 95.80\% for the BF16 reference
(\cref{app:oft}). On GR00T N1.7 evaluated on LIBERO-Pro~\citep{liberopro2025}, SACQ
achieves 43.62\% at 2 L/A bpw, compared with 43.67\% for BF16
(\cref{app:libero_pro}).
These results indicate that SACQ remains effective across different VLA
architectures and under a broader range of perturbations.
Validation on real robotic systems remains an important direction for future work.

\bibliography{references}
\bibliographystyle{iclr2027_conference}

\appendix
\crefalias{section}{appendix}
\crefalias{subsection}{appendix}
\crefalias{subsubsection}{appendix}

\section{Structure-Aware Codebook Fitting Details}
\label{app:structure}

\subsection{Metric-Weighted Codebook Optimization}
\label{app:weighted_codebook_optimization}

As summarized in \cref{sec:codebook-opt}, given $G(v)$ and $\alpha(v)$, we optimize
the objective in \cref{eq:metric-kmeans} by alternating between
two steps, as in $k$-means. With the codewords fixed, each vector is assigned
to the codeword nearest under its own metric,
\begin{equation}
a(v) = \arg\min_j \;
(v-c_j)^{\!\top} G(v) (v-c_j),
\label{eq:assign}
\end{equation}
and with the assignments fixed, each codeword is the solution of a weighted
least-squares problem over the vectors assigned to it,

\begin{equation}
c_j =
\left(
\sum_{v:a(v)=j} \alpha(v)G(v)
\right)^+
\sum_{v:a(v)=j} \alpha(v)G(v)v.
\label{eq:update}
\end{equation}

where $(\cdot)^+$ denotes the pseudoinverse. Under these exact, unconstrained updates, both steps are non-increasing in \cref{eq:metric-kmeans}.
\subsection{Language-Backbone Feed-Forward Metric Derivation}
\label{app:backbone_metrics}

We provide the first-order derivation underlying the feed-forward metrics
summarized in \cref{sec:structure_metric}.

The feed-forward block is given by
\[
g = W_g h,\qquad
u = W_u h,\qquad
z = \operatorname{SiLU}(g)\odot u,\qquad
y = W_d z.
\]
Here, $z$ is the input to the down projection and $y$ is the FFN output.
For the down projection, the input-side Gram is
$A=\mathbb{E}[zz^\top]$, and we use $B=I$ as for the output projection.

The gate and up projections require different treatment because their outputs
interact nonlinearly through the SwiGLU operation before being passed through
the down projection $W_d$. We therefore approximate their local perturbations
to first order. Consider a four-dimensional weight vector $v$ in output
channel $r$, and let $h_v\in\mathbb{R}^4$ denote the input activations
corresponding to its four weights. For the gate projection, the change in
$z_r$ under a small perturbation $\delta v$ is approximated as
\[
\delta z_r
\approx
\operatorname{SiLU}'(g_r)\,u_r\,h_v^\top\delta v.
\]

This change propagates through the down projection to the FFN output as
$\delta y = W_d[:,r]\,\delta z_r$,
and therefore
\[
\|\delta y\|_2^2
=
\|W_d[:,r]\|_2^2(\delta z_r)^2.
\]
Substituting the first-order approximation above and taking the expectation
over calibration samples gives the local metrics
\begin{equation}
\begin{aligned}
G_{\mathrm{gate}}(v)
&=
\|W_d[:,r]\|_2^2
\mathbb{E}\!\left[
\bigl(\operatorname{SiLU}'(g_r)u_r\bigr)^2
h_vh_v^\top
\right],\\
G_{\mathrm{up}}(v)
&=
\|W_d[:,r]\|_2^2
\mathbb{E}\!\left[
\operatorname{SiLU}(g_r)^2
h_vh_v^\top
\right].
\end{aligned}
\end{equation}

In practice, we retain only the diagonal entries of these $4\times4$ local
metrics to simplify codebook fitting:
\[
\widetilde{G}(v)
=
\operatorname{Diag}\!\left(
G_{11}(v), G_{22}(v), G_{33}(v), G_{44}(v)
\right),
\]
where $G(v)$ denotes the corresponding gate or up metric above.
This approximation retains the downstream sensitivity induced by the down
projection while discarding cross-coordinate correlations within each weight
vector.

\subsection{Action-Head Metric Derivations}
\label{app:action_head_metrics}

The projection roles in the DiT action head follow the
structure-aware construction in \cref{sec:structure_metric}, with three
model-specific differences.

First, the DiT action head contains both self-attention and cross-attention
blocks. In cross-attention, the query-side input does not coincide with the
key/value-side input. We therefore collect the two separately, with Grams
\[
A_q = \mathbb{E}[x_qx_q^\top],
\qquad
A_{kv} = \mathbb{E}[x_{kv}x_{kv}^\top].
\]
The query projection uses $A_q$ as its input-side Gram and the key
projection uses $A_{kv}$; in self-attention the two coincide.
The output-side weighting is obtained from the second moment of the
opposite projection, namely
$\operatorname{diag}(W_K A_{kv} W_K^\top)$ for the query projection and
$\operatorname{diag}(W_Q A_q W_Q^\top)$ for the key projection.
For grouped-query attention, each key-side energy is repeated for the query
heads mapped to that key/value head, whereas query-side energies are summed
over the query heads sharing the same key/value head.
The value projection uses the key/value-side inputs with an input-side Gram
weighted by the squared attention probabilities, following the construction
in \cref{sec:structure_metric}.

Second, the feed-forward block uses a GELU activation rather than the SwiGLU
structure of the language backbone:
\[
p = W_{\mathrm{up}}h,\qquad
z = \operatorname{GELU}(p),\qquad
y = W_{\mathrm{down}}z.
\]
Consider a four-dimensional weight vector $v$ in output channel $r$, and let
$h_v$ denote the input activations corresponding to its four weights.
Under a small perturbation $\delta v$,
\[
\delta p_r = h_v^\top\delta v,
\qquad
\delta z_r
\approx
\operatorname{GELU}'(p_r)h_v^\top\delta v.
\]
This change propagates through the down projection as
$\delta y=W_{\mathrm{down}}[:,r]\,\delta z_r$,
and therefore
\[
\|\delta y\|_2^2
\approx
\|W_{\mathrm{down}}[:,r]\|_2^2
\operatorname{GELU}'(p_r)^2
(h_v^\top\delta v)^2.
\]

In our implementation, we use the separable approximation
\[
\mathbb{E}\!\left[
\operatorname{GELU}'(p_r)^2 h_vh_v^\top
\right]
\approx
\mathbb{E}\!\left[\operatorname{GELU}'(p_r)^2\right]
\mathbb{E}\!\left[h_vh_v^\top\right].
\]
The metric used for codebook fitting is therefore
\begin{equation}
G_{\mathrm{up}}(v)
=
\|W_{\mathrm{down}}[:,r]\|_2^2
\mathbb{E}\!\left[
\mathrm{GELU}'(p_r)^2
\right]
\mathbb{E}\!\left[
h_v h_v^\top
\right].
\end{equation}
This approximation discards correlations between the activation sensitivity
and input covariance. The down projection uses the Gram of its post-activation
input $z$ with $B=I$, as in the language backbone.

Third, the adaptive layer normalization projection requires its own
output-side weighting. Let $t_{\mathrm{emb}}$ denote the diffusion timestep
embedding. The projection receives the post-activation conditioning input
$c=\mathrm{SiLU}(t_{\mathrm{emb}})$ and produces the modulation parameters
in (scale, shift) order, which are applied as
\[
y
=
\operatorname{LayerNorm}(x)\odot(1+\mathrm{scale})
+\mathrm{shift}.
\]
An error in the scale branch is multiplied by the normalized hidden state as it
propagates, whereas an error in the shift branch propagates with unit
coefficient. Writing
$\nu_j=\mathbb{E}[\operatorname{LayerNorm}(x)_j^2]$,
we use the separable approximation
\[
    \mathbb{E}\!\left[
        \operatorname{LayerNorm}(x)_j^2 cc^\top
    \right]
    \approx
    \mathbb{E}\!\left[
        \operatorname{LayerNorm}(x)_j^2
    \right]
    \mathbb{E}\!\left[cc^\top\right],
\]
estimating the two factors separately.
The output-side weighting is $[\nu; \mathbf{1}]$ and the input-side Gram is
$A = \mathbb{E}[cc^\top]$.

\subsection{Complete Metric Definitions}
\label{app:metric_summary}

For completeness, we summarize the final structure-aware metrics used for
all projection roles in the language backbone and DiT action head.

\begin{table}[H]
    \centering
    \small
    \renewcommand{\arraystretch}{1.22}
    \setlength{\tabcolsep}{5pt}
    \caption{\textbf{Metric construction for the seven language backbone roles.}
    Head and group indices are omitted for readability; the actual computation
    applies the corresponding normalization, RoPE, and GQA repetition or
    aggregation.}
    \label{tab:vlm_metric}

    \begin{tabular*}{\linewidth}{
        @{\extracolsep{\fill}}
        lcc
        @{}
    }
        \toprule
        Projection
        & Input-side Gram $A$
        & Output-side weighting $B_{rr}$ \\
        \midrule

        q proj
        & $\mathbb{E}[xx^\top]$
        & $\mathbb{E}[K_r^2]$ \\

        k proj
        & $\mathbb{E}[xx^\top]$
        & $\mathbb{E}[Q_r^2]$ \\

        v proj
        & $\displaystyle
          \mathbb{E}\!\left[
          \sum\nolimits_s
          \omega_s x_sx_s^\top
          \right],
          \quad
          \omega_s=
          \sum\nolimits_{h,t}P_{h,t,s}^2$
        & $\|W_O[:,r]\|_2^2$ \\

        o proj
        & $\mathbb{E}[C^\top C],\quad C=PV$
        & $1$ \\

        down proj
        & $\displaystyle
          \mathbb{E}[zz^\top],
          \quad
          z=\mathrm{SiLU}(g)\odot u$
        & $1$ \\

        \midrule
        \multicolumn{3}{l}{\textit{Direct diagonal metric}} \\

        gate proj
        & \multicolumn{2}{c}{
          $\displaystyle
          G(v)
          =
          \|W_d[:,r]\|_2^2
          \mathrm{Diag}\!\left(
          \mathbb{E}\!\left[
          (\mathrm{SiLU}'(g_r)u_r)^2
          (h_v\odot h_v)
          \right]
          \right)$
        } \\

        up proj
        & \multicolumn{2}{c}{
          $\displaystyle
          G(v)
          =
          \|W_d[:,r]\|_2^2
          \mathrm{Diag}\!\left(
          \mathbb{E}\!\left[
          \mathrm{SiLU}(g_r)^2
          (h_v\odot h_v)
          \right]
          \right)$
        } \\

        \bottomrule
    \end{tabular*}
\end{table}

\begin{table}[H]
    \centering
    \small
    \renewcommand{\arraystretch}{1.22}
    \setlength{\tabcolsep}{5pt}
    \caption{\textbf{Metric construction for the seven action head roles.}
    Here, $x_q$ and $x_{kv}$ are the query-side and key/value-side attention
    inputs, which coincide in self-attention.
    The two output projections, proj\_out\_1 and proj\_out\_2, are not among these roles and use Euclidean
    vector k-means.}
    \label{tab:action_metric}

    \begin{tabular}{lcc}
        \toprule
        Projection
        & Input-side Gram $A$
        & Output-side weighting $B_{rr}$ \\
        \midrule

        Q
        & $A_q=\mathbb{E}[x_qx_q^\top]$
        & $\mathrm{diag}(W_K A_{kv} W_K^\top)_r$ \\

        K
        & $A_{kv}=\mathbb{E}[x_{kv}x_{kv}^\top]$
        & $\mathrm{diag}(W_Q A_q W_Q^\top)_r$ \\

        V
        & $\displaystyle
          \mathbb{E}\!\left[
          \sum\nolimits_s
          \omega_s x_{kv,s}x_{kv,s}^\top
          \right],
          \quad
          \omega_s=
          \sum\nolimits_{h,t}P_{h,t,s}^2$
        & $\|W_O[:,r]\|_2^2$ \\

        O
        & $\mathbb{E}[C^\top C],\quad C=PV$
        & $1$ \\

        FF-up
        & $\mathbb{E}[hh^\top]$
        & $\displaystyle
          \mathbb{E}\!\left[
          \mathrm{GELU}'\!\left(
          (W_{\mathrm{up}}h)_r
          \right)^2
          \right]
          \|W_{\mathrm{down}}[:,r]\|_2^2$ \\

        FF-down
        & $\displaystyle
          \mathbb{E}[zz^\top],
          \quad
          z=\mathrm{GELU}(W_{\mathrm{up}}h)$
        & $1$ \\

        AdaLN
        & $\displaystyle
          \mathbb{E}[cc^\top],
          \quad
          c=\mathrm{SiLU}(t_{\mathrm{emb}})$
        & $\displaystyle
          [\nu;\mathbf{1}]_r,
          \quad
          \nu_j=
          \mathbb{E}\!\left[
          \mathrm{LayerNorm}(x)_j^2
          \right]$ \\

        \bottomrule
    \end{tabular}
\end{table}

\subsection{Fitting Configuration}
\label{app:fitting_configuration}

All codebook fitting is performed on the scale-normalized weight vectors defined in \cref{sec:vc}, with $\alpha(v)=1$ as described in \cref{sec:structure_metric}.
For each layer, we fit the codebook on a
sampled subset of weight vectors and, once the codebook is fixed, recompute
the assignments over all vectors in that layer.

We use different fitting budgets for the language backbone and DiT action
head. The language-backbone codebooks use at most 32 calibration requests,
with up to 2,048 cached tokens per layer, 50,000 sampled weight vectors, and
8 Lloyd iterations. The DiT codebooks use 32 calibration requests per suite,
200,000 sampled weight vectors, and 12 Lloyd iterations.

Codewords are initialized from sampled weight vectors. For full-metric codeword updates,
we add an unscaled ridge term of $10^{-4}I$ to the normal matrix in \cref{eq:update}
for numerical stability. Updated codewords are clipped to $[-1,1]$.

\section{Action-Aware Mixed-Bit Allocation Details}
\label{app:allocation}

\subsection{Iterative Allocation Procedure}
\label{app:allocation_procedure}

\begin{algorithm}[H]
\caption{Action-aware mixed-bit allocation}
\label{alg:mixed_bit_allocation}
\begin{algorithmic}[1]
\STATE Initialize every candidate layer at $\mathrm{GPTQ\,(4bit)}$.
\WHILE{the calculated bits per weight exceed the target budget}
    \STATE Periodically measure the one-step action-error increases
    $\Delta_\ell$, averaged over the calibration set with equal weighting
    across suites.
    \STATE Compute the calculated bits per weight of each component.
    \STATE Select the component with the higher calculated bits per weight.
    \STATE Within it, choose $\ell^\star$ by
    \cref{eq:allocation_rule}.
    \STATE Move $\ell^\star$ to its next lower-precision state.
\ENDWHILE
\end{algorithmic}
\end{algorithm}

The measurements $\Delta_\ell$ are refreshed after a fixed number of accepted
moves, making the procedure a periodically refreshed greedy search rather than
a globally optimal allocation.
\subsection{Allocation Configuration}
\label{app:allocation_configuration}

For LIBERO, we evaluate allocation sensitivity using 10 fixed calibration requests from
each of the four LIBERO suites, giving 40 requests in total. The BF16 reference
action for each request is generated using the corresponding suite-specific
checkpoint. Let $\mathbf{a}_{Q,i}$ denote the final decoded action tensor produced by the
quantized policy for request $i$, and let
$\mathbf{a}_{\mathrm{BF16},i}$ denote its BF16 reference.
We define the request-level relative action error as
\[
e_i =
\frac{
\left\|\mathbf{a}_{Q,i}-\mathbf{a}_{\mathrm{BF16},i}\right\|_F
}{
\max\!\left(
\left\|\mathbf{a}_{\mathrm{BF16},i}\right\|_F,
10^{-9}
\right)
}.
\]
The norm is taken over the full decoded action tensor of each request. The
current allocation error is 
$E=\frac{1}{40}\sum_{i=1}^{40}e_i$.
Since each suite contributes the same number of requests, this is equivalent
to equally weighting the four suites.

Let $E_\ell^{\downarrow}$ denote the average relative error when only layer
$\ell$ is moved to its next lower-precision state. We define its sensitivity as
\begin{equation}
\Delta_\ell = E_\ell^{\downarrow} - E .
\end{equation}
Thus, $\Delta_\ell$ measures the increase in final decoded-action relative error caused
by a one-step precision reduction of layer $\ell$. We use the same policy-noise
seed when computing these action errors.

For SimplER Google, we use the same relative-error definition and evaluate
allocation sensitivities on 64 fixed requests, with eight requests from each
of the four task categories under each evaluation mode, Visual Matching and
Variant Aggregation. These requests are shared with codebook fitting, and
request-level relative action errors are averaged uniformly.

Sensitivities are refreshed every 100 accepted moves. Calculated bpw is
used for allocation cost and component balancing, while checkpoints are
saved according to nominal bpw for convenient comparison with other methods.
Candidate layers transition only between adjacent quantization states.

\section{Post-Quantization Refinement Details}
\label{app:refinement}
\subsection{Second-Order Index Reassignment}
\label{app:index_reassignment}

For each LIBERO suite, we use all 576 training requests to collect
layer-input statistics, excluding the 64 validation requests. With the
quantized policy frozen after structure-aware fitting and mixed-bit
allocation, we collect the full input second-moment matrix $H_\ell$
from all input rows of each target layer. For the DiT action head,
inputs from all denoising calls are included.

For numerical stability, we use
\begin{equation}
\widetilde H_\ell
=
\frac{1}{2}(H_\ell+H_\ell^\top)
+
0.1\,
\frac{\operatorname{tr}(H_\ell)}{n_\ell}I,
\qquad
\widetilde H_\ell^{-1}=U_\ell^\top U_\ell ,
\end{equation}
where $n_\ell$ denotes the input dimension of layer $\ell$.
The reconstruction target is the original BF16 weight. During index
reassignment, weight vectors are processed sequentially, while the
codebooks, row-group scales, mixed-bit allocation, and all other model
parameters remain fixed. Only the codeword indices are updated.

\subsection{Action-Level Calibration}
\label{app:action_calibration}

After index reassignment, we freeze the updated codeword indices and
optimize only the codebooks and row-group scales. As described in
\cref{sec:refinement}, training minimizes the MSE between the final
decoded actions of the quantized policy and the BF16 reference. For stochastic action generation, we use the same policy-noise seed for
the quantized and BF16 policies when computing the action loss.

We use the same calibration configuration for GR00T N1.7 and OpenVLA-OFT on LIBERO. We use AdamW with codebook and scale learning rates of
$10^{-4}$ and $10^{-5}$, respectively, zero weight decay, and no
learning-rate scheduler or warm-up. For each suite, we use 576 training
requests and 64 validation requests with a global batch size of 6.
Calibration runs for 12 epochs, corresponding to 1,152 optimizer updates,
with gradient-norm clipping at 1.0.

For SimplER Google, we use the same optimization procedure with codebook
and scale learning rates of $5\times10^{-5}$ and $5\times10^{-6}$,
respectively. We use 512 training requests and 128 validation requests
with a global batch size of 8, and train for 12 epochs, corresponding to
768 optimizer updates. In all settings, only the codebooks and scales are
optimized; the updated indices and all other model parameters remain fixed.

We select the checkpoint with the lowest mean relative action error on the
validation set,
\[
P_{\mathrm{val}}
=
\frac{1}{|\mathcal{D}_{\mathrm{val}}|}
\sum_{i\in\mathcal{D}_{\mathrm{val}}}
\frac{\|a_{Q,i}-a_{\mathrm{BF16},i}\|_F}
{\max(\|a_{\mathrm{BF16},i}\|_F,10^{-9})}.
\]
Here, $\mathcal{D}_{\mathrm{val}}$ denotes the validation set.

\section{Experimental and Quantization Details}
\label{app:experimental_details}
\subsection{Evaluation Protocols}
\label{app:evaluation_protocols}

We use LIBERO~\citep{libero2023} and SimplER~\citep{li24simpler}
to evaluate the quantized policies and follow the official evaluation
protocols and action-execution settings.

For LIBERO, we evaluate the Spatial, Object, Goal, and Long suites, each
consisting of 10 tasks with 50 official initial states per task, resulting
in 500 episodes per suite and 2,000 episodes per model configuration.
All evaluations use environment seed 17, and at each policy inference step
we execute the first eight actions of the predicted action chunk.

For SimplER, we evaluate the Google Robot embodiment using the Visual
Matching (VM) and Variant Aggregation (VA) protocols. VM comprises 864
evaluation conditions and VA 1,992, with both protocols covering four task
categories: Pick, Move, Drawer, and Apple. All evaluations use seed 7, and
at each policy inference step we execute only the first action of the
predicted action chunk.

\subsection{Quantization Scope and Bit Reporting}
\label{app:quantization_scope}

For GR00T N1.7, the quantization candidate set contains 338 linear
layers: 112 in the language backbone and 226 in the DiT action head.
Outside this candidate scope, 128 additional linear layers and the token
embedding are fixed at 4-bit precision in the quantized models. Of these
128 linear layers, 104 belong to the vision encoder/projector. We observed that the GR00T N1.7 vision encoder is sensitive to sub-4-bit quantization and therefore we keep it at 4-bit.
The vision encoder is quantized to 4-bit using GPTQ, while the remaining
fixed components use 4-bit affine quantization. 
Except for the A8 inference setting in \cref{sec:inference_efficiency},
activations remain BF16, and BF16 serves as the full-precision reference.

All baselines are applied to the same language/action candidate scope.
The L/A bpw reported in the main text and tables denotes the
parameter-weighted average bits per weight over these 338 candidate
layers and excludes metadata such as codebooks and scales. In contrast,
the reported memory footprint is measured for the full loaded policy.
\section{Baseline Implementation Details}
\label{app:baseline_details}

\paragraph{HQQ.}
We apply uniform HQQ to all 338 language/action candidate layers with
group size 64, evaluating W3 and W2 configurations. Activations and
computation remain in BF16.

\paragraph{GPTQ.}
We apply GPTQ to the same 338 candidate layers with group size 64 and
block size 128. For an additional ultra-low-bit comparison, we extend
the same GPTQ port to symmetric ternary codes $\{-1,0,+1\}$ with
group-64 scaling and GPTQ-style error compensation. This setting has a
nominal rate of $\log_2(3)\approx1.585$ bpw and is a custom ternary
extension rather than a standard GPTQ mode.

\paragraph{ActQuant.}
We apply ActQuant's inter-tensor bit allocation and intra-tensor scale
optimization to the same 338 language/action candidate layers. Since
ActQuant does not directly target GR00T's DiT action head, we apply
its procedure to the DiT layers to match the quantization scope used
in our comparisons.

\paragraph{TWLA.}
We port TWLA to all 338 candidate layers
while keeping activations in BF16. However, our PyTorch VLA port does
not directly reproduce the compact llama.cpp inference path and some
of the storage/runtime optimizations described in the original work.
Consequently, the resident memory reported for our TWLA port may be
higher than what a more optimized TWLA implementation could achieve.
In addition, one language-backbone projection produces non-finite
values during quantization, for which we use an FP64 fallback.

\paragraph{AQLM.}
We apply the official AQLM implementation to the same 338 candidate layers. We use group size 8 with
$1{\times}12$, $1{\times}16$, and $2{\times}12$ codebook configurations
for nominal 1.5-, 2-, and 3-bit budgets, respectively. The 2-bit and 3-bit
configurations follow settings reported by AQLM, while $1{\times}12$
extends AQLM to the 1.5-bit setting.
Nominal bpw excludes codebook and scale metadata, while the reported
resident memory includes these costs. The 2-bit $1{\times}16$
configuration uses a 65,536-entry codebook per layer, resulting in
larger codebook overhead than the 3-bit $2{\times}12$ configuration and slightly higher measured resident memory despite the lower
nominal bit rate.

\section{Additional Experiments}
\label{app:additional_experiments}

We conduct two additional evaluations beyond the main experiments.
First, we apply SACQ to OpenVLA-OFT and evaluate the quantized model
on LIBERO to examine whether its low-bit behavior extends to a
different VLA architecture.
Second, we evaluate GR00T N1.7 quantized with SACQ on LIBERO-Pro to examine whether the quantized policy remains close to its BF16 reference under a broader range of perturbations.

\subsection{Cross-Architecture Evaluation on OpenVLA-OFT}
\label{app:oft}

We evaluate OpenVLA-OFT on the four LIBERO suites using 50 official
initial states per task, resulting in 500 episodes per suite. We use
program and policy seed 17, environment seed 17, and execute an action
chunk of length 8. Unlike the GR00T setting, the quantization scope
contains 420 linear layers: 196 in the vision encoder and 224 in the
language backbone, while the action head is excluded. 
We also apply structure-aware fitting to the vision-encoder linear layers,
while keeping the remaining SACQ procedure unchanged.
We therefore
report V/L bpw for OpenVLA-OFT, which is not directly comparable to the
L/A bpw reported for GR00T in the main experiments.

\begin{table}[H]
    \centering
    \small
    \renewcommand{\arraystretch}{1.05}
    \caption{
    \textbf{Cross-architecture evaluation on OpenVLA-OFT.}
    Success rates (\%) are reported across the four LIBERO suites.
    V/L bpw denotes the quantization rate over the 420 vision and
    language linear layers.
    }
    \label{tab:oft}

    \begin{tabular*}{\linewidth}{
        @{\extracolsep{\fill}}
        lccccccc
        @{}
    }
        \toprule
        Method & V/L bpw & Spatial & Object & Goal & Long & Avg. & Memory (GB) \\
        \midrule

        BF16
        & 16.00
        & 93.8 & 98.2 & 97.2 & 94.0
        & 95.80 & 15.203 \\

        \midrule

        \textbf{SACQ}
        & 3.00
        & 91.8 & 97.4 & 97.8 & 95.2
        & 95.55 & 5.128 \\

        \midrule

        GPTQ
        & 2.00
        & 67.6 & 55.8 & 73.6 & 49.6
        & 61.65 & 3.625 \\

        ActQuant
        & 2.00
        & 79.6 & 60.8 & 77.2 & 24.8
        & 60.60 & 2.953 \\

        TWLA
        & 2.11
        & 91.0 & \textbf{96.4} & 96.2
        & \textbf{93.8}
        & \textbf{94.35} & 8.960 \\

        \textbf{SACQ}
        & 2.00
        & \textbf{92.0} & 96.0 & \textbf{98.0}
        & 90.2
        & 94.05 & 2.926 \\

        \midrule

        GPTQ-Ternary
        & 1.58
        & 0.0 & 0.0 & 0.0 & 0.0
        & 0.00 & 3.026 \\

        ActQuant
        & 1.50
        & 61.6 & 31.2 & 35.2 & 0.0
        & 32.00 & 2.433 \\

        \textbf{SACQ}
        & 1.50
        & \textbf{89.4} & \textbf{95.0} & \textbf{92.8}
        & \textbf{65.6}
        & \textbf{85.70} & 2.463 \\

        \bottomrule
    \end{tabular*}
\end{table}

As shown in \cref{tab:oft}, SACQ achieves 95.55\% average success
at 3 V/L bpw, only 0.25 percentage points below the 95.80\% BF16
reference. At 2 V/L bpw, SACQ retains 94.05\% average success, similar
to the 94.35\% obtained by TWLA and substantially higher than the
evaluated GPTQ and ActQuant variants. These results show that the
low-bit quantization performance of SACQ extends beyond GR00T N1.7
to a different VLA architecture.

\subsection{Additional Evaluation on LIBERO-Pro}
\label{app:libero_pro}

We additionally evaluate SACQ on LIBERO-Pro~\citep{liberopro2025} using GR00T N1.7.
LIBERO-Pro includes five perturbation families: Language, Object, Swap,
Task, and Environment. Across the four LIBERO suites, each model is
evaluated over 10,000 episodes in total.

\begin{table}[H]
    \centering
    \caption{\textbf{Evaluation of GR00T N1.7 on LIBERO-Pro.}
    Success rates (\%) are aggregated over the five perturbation
    families for each LIBERO suite.}
    \label{tab:libero_pro}
    \small
    \renewcommand{\arraystretch}{1.05}

    \begin{tabular*}{\linewidth}{
        @{\extracolsep{\fill}}
        lcccccc
        @{}
    }
        \toprule
        Method & L/A bpw & Spatial & Object & Goal & Long & Avg. \\
        \midrule

        BF16
        & 16.0
        & 44.60 & 41.28 & 45.52 & 43.28
        & 43.67 \\

        SACQ
        & 3.00
        & 44.44 & 41.04 & 45.56 & 44.00
        & 43.76 \\

        SACQ
        & 2.00
        & 44.60 & 41.08 & 44.48 & 44.32
        & 43.62 \\

        SACQ
        & 1.50
        & 43.64 & 40.72 & 43.24 & 42.88
        & 42.62 \\

        \bottomrule
    \end{tabular*}
\end{table}

As shown in \cref{tab:libero_pro}, BF16 achieves 43.67\% average success,
while SACQ achieves 43.76\%, 43.62\%, and 42.62\% at 3, 2, and 1.5 L/A bpw,
respectively.
The 3-bit and 2-bit models remain nearly identical to the BF16 reference
in overall success, while the 1.5-bit model also maintains competitive
performance.
These results show that SACQ maintains performance close to the BF16
reference even under a range of perturbations.

\section{Structure-Aware Fitting under Limited Refinement}
\label{app:limited_refinement}

We further examine the role of structure-aware codebook fitting when
post-quantization refinement is performed under a limited refinement budget. At 1.5 L/A bpw, we compare three codebook-fitting
objectives while keeping the mixed-bit allocation fixed:
structure-aware fitting, layer-local reconstruction fitting, and
Euclidean fitting. 
Each training request yields five calibration observations, so the full
set of 576 training requests corresponds to 2,880 observations per suite.
The layer-local baseline fits codebooks according
to the reconstruction error at the output of each individual projection,
\[
\mathcal{L}_{\mathrm{local}}
=
\left\|
X(W-\widehat{W})^\top
\right\|_F^2,
\]
without modeling the downstream operation in which that projection
output is subsequently used. This layer-local objective follows the
reconstruction criterion commonly used in prior vector-quantization
methods such as GPTVQ~\citep{gptvq2024}. The remaining
quantization and refinement pipeline is kept identical across all
settings, and refinement is performed for 12 epochs.

\begin{table}[t]
    \centering
    \caption{\textbf{Codebook fitting under limited post-quantization
    refinement at 1.5 L/A bpw.}
    We vary the calibration data per LIBERO suite while fixing
    refinement to 12 epochs. $\Delta$ denotes the difference in
    four-suite average success relative to structure-aware fitting
    under the same calibration budget.}
    \label{tab:limited_refinement}
    \small
    \renewcommand{\arraystretch}{1.05}
    \begin{tabular*}{\textwidth}{
        @{\extracolsep{\fill}}
        r l r r r r r r
        @{}
    }
        \toprule
        Calib. data &
        Fitting &
        Spatial &
        Object &
        Goal &
        Long &
        Avg. &
        $\Delta$ vs. Struct. \\
        \midrule

        --
        & BF16
        & 94.2
        & 97.6
        & 95.2
        & 89.4
        & 94.10
        & -- \\
        \midrule

        90
        & Structure-aware
        & \textbf{76.2}
        & 94.6
        & \textbf{88.0}
        & \textbf{77.8}
        & \textbf{84.15}
        & 0.00 \\
        & Layer-local
        & 61.8
        & \textbf{95.0}
        & 71.8
        & 69.6
        & 74.55
        & -9.60 \\
        & Euclidean
        & 42.2
        & 48.6
        & 35.0
        & 33.2
        & 39.75
        & -44.40 \\
        \midrule

        180
        & Structure-aware
        & \textbf{86.0}
        & 95.8
        & \textbf{90.6}
        & \textbf{83.0}
        & \textbf{88.85}
        & 0.00 \\
        & Layer-local
        & 78.8
        & \textbf{96.4}
        & 87.6
        & 73.4
        & 84.05
        & -4.80 \\
        & Euclidean
        & 48.2
        & 87.4
        & 52.0
        & 41.0
        & 57.15
        & -31.70 \\
        \midrule

        360
        & Structure-aware
        & \textbf{91.2}
        & 96.6
        & \textbf{94.0}
        & \textbf{81.0}
        & \textbf{90.70}
        & 0.00 \\
        & Layer-local
        & 87.0
        & \textbf{97.2}
        & 88.2
        & 80.0
        & 88.10
        & -2.60 \\
        & Euclidean
        & 78.4
        & 95.2
        & 77.2
        & 63.4
        & 78.55
        & -12.15 \\
        \midrule

        720
        & Structure-aware
        & \textbf{91.8}
        & \textbf{96.2}
        & 93.0
        & 84.0
        & 91.25
        & 0.00 \\
        & Layer-local
        & 90.6
        & 95.8
        & \textbf{94.2}
        & \textbf{85.2}
        & \textbf{91.45}
        & +0.20 \\
        & Euclidean
        & 87.8
        & 94.2
        & 88.4
        & 75.4
        & 86.45
        & -4.80 \\
        \midrule

        1440
        & Structure-aware
        & 93.2
        & \textbf{97.6}
        & 93.8
        & 85.6
        & 92.55
        & 0.00 \\
        & Layer-local
        & 92.6
        & 97.4
        & \textbf{95.0}
        & \textbf{86.8}
        & \textbf{92.95}
        & +0.40 \\
        & Euclidean
        & \textbf{93.6}
        & \textbf{97.6}
        & 91.6
        & 79.6
        & 90.60
        & -1.95 \\
        \midrule

        2880
        & Structure-aware
        & 93.2
        & \textbf{97.0}
        & \textbf{96.2}
        & \textbf{89.0}
        & \textbf{93.85}
        & 0.00 \\
        & Layer-local
        & 92.6
        & \textbf{97.0}
        & 95.0
        & 85.8
        & 92.60
        & -1.25 \\
        & Euclidean
        & \textbf{95.2}
        & 95.8
        & 93.8
        & 85.6
        & 92.60
        & -1.25 \\
        \bottomrule
    \end{tabular*}
\end{table}

As shown in \cref{tab:limited_refinement}, differences among the
codebook-fitting objectives are most pronounced at smaller calibration
budgets. With only 90 calibration observations per suite,
structure-aware fitting achieves 84.15\% average success, compared with
74.55\% for layer-local reconstruction fitting and 39.75\% for
Euclidean fitting. These differences narrow rapidly as the calibration
data increase. At 720 and 1,440 observations, layer-local reconstruction
fitting achieves averages only 0.20 and 0.40 percentage points higher
than structure-aware fitting, respectively, leaving the two methods
numerically close. With the full 2,880 observations, structure-aware
fitting achieves the highest average success of 93.85\%. On
LIBERO-Long, it recovers to 89.0\%, close to the 89.4\% BF16 reference,
whereas layer-local reconstruction and Euclidean fitting reach 85.8\%
and 85.6\%, respectively.

Because refinement is fixed to 12 epochs in all settings, reducing the
calibration data also reduces both the number of optimization updates
and the amount of input statistics available for index reassignment.
Under this setting, the results indicate that the main benefit of
structure-aware fitting is not uniformly higher final average
performance at every calibration budget, but a stronger low-bit
initialization when subsequent refinement is limited. As the calibration
budget increases, the differences in overall average success become
smaller; however, in the full-budget setting, structure-aware fitting
recovers LIBERO-Long to 89.0\%, closest to the 89.4\% BF16 reference.
This suggests that even when the overall averages become similar,
structure-aware fitting can preserve LIBERO-Long performance closer to the
BF16 reference.

\end{document}